\documentclass{article}
\usepackage{ijcai23}

\usepackage{times}
\usepackage{soul}
\usepackage{url}
\usepackage[hidelinks]{hyperref}
\usepackage[utf8]{inputenc}
\usepackage[small]{caption}
\usepackage{graphicx}
\usepackage{multirow}
\usepackage{amsmath}
\usepackage{amsthm}
\usepackage{amsfonts}
\usepackage{xcolor}
\usepackage{subfig}
\usepackage{amsmath}
\usepackage{booktabs}
\usepackage{algorithm}
\usepackage{algorithmic}
\usepackage[normalem]{ulem}
\usepackage[switch]{lineno}
\DeclareMathOperator*{\argmax}{argmax}

\title{HGAttack: Transferable Heterogeneous Graph Adversarial Attack}

\author{
He Zhao$^1$
\and
Zhiwei Zeng$^1$\and
Yongwei Wang$^{1}$\and
Deheng Ye$^{2}$\And
Chunyan Miao$^1$
\affiliations
$^1$Nanyang Technological University\\
$^2$Tencent
\emails
ZHAO0378@e.ntu.edu.sg,
\{zhiwei.zeng, ascymiao\}@ntu.edu.sg,
yongweiw@outlook.com,
dericye@tencent.com 
}

\begin{document}

\maketitle

\begin{abstract}
Heterogeneous Graph Neural Networks (HGNNs) are increasingly recognized for their performance in areas like the web and e-commerce, where resilience against adversarial attacks is crucial. However, existing adversarial attack methods, which are primarily designed for homogeneous graphs, fall short when applied to HGNNs due to their limited ability to address the structural and semantic complexity of HGNNs. This paper introduces HGAttack, the first dedicated gray box evasion attack method for heterogeneous graphs. We design a novel surrogate model to closely resemble the behaviors of the target HGNN and utilize gradient-based methods for perturbation generation. Specifically, the proposed surrogate model effectively leverages heterogeneous information by extracting meta-path induced subgraphs and applying GNNs to learn node embeddings with distinct semantics from each subgraph. This approach improves the transferability of generated attacks on the target HGNN and significantly reduces memory costs. For perturbation generation, we introduce a semantics-aware mechanism that leverages subgraph gradient information to autonomously identify vulnerable edges across a wide range of relations within a constrained perturbation budget. We validate HGAttack's efficacy with comprehensive experiments on three datasets, providing empirical analyses of its generated perturbations. Outperforming baseline methods, HGAttack demonstrated significant efficacy in diminishing the performance of target HGNN models, affirming the effectiveness of our approach in evaluating the robustness of HGNNs against adversarial attacks.


\end{abstract}

\section{Introduction}

Many real-world data naturally possess a heterogeneous graph (HG) structure. Heterogeneous Graph Neural Networks (HGNNs) have shown promising performance in extracting diverse and complex heterogeneous information in various domains, such as the web ~\cite{zhang2019key} and e-commerce~\cite{ji2022prohibited,lv2021we}.
However, the resilience of HGNNs against adversarial attacks, which frequently occur in these domains, has not been thoroughly explored. Given the increasing reliance on HGNNs in critical applications, understanding and improving their robustness against such attacks is imperative. To this end, there is a need to develop
a heterogeneous graph adversarial attack method, to thoroughly evaluate the vulnerabilities of HGNNs and ensure their robustness.

Most current graph adversarial methods are designed for homogeneous graphs~\cite{zugner2018adversarial,chen2018fast,li2022revisiting,sharma2023task}. However, when it comes to the case of heterogeneous graphs, homogeneous adversarial attack methods tend to underperform, primarily due to their limited ability to accommodate the structural complexities and diverse semantic relations inherent in these graphs.

In this paper, we aim to propose an adversarial attack method for heterogeneous graphs in the context of \textit{Targeted Evasion Gray Box Attack}. 
In this scenario, attackers are limited to accessing only the training labels and inputs, i.e. adjacency matrices and feature matrices, while remaining uninformed about the underlying structure of the target model. 
Prevalent gray box attacks employ a surrogate model such as GCN~\cite{kipf2016semi} and apply gradient-based methods~\cite{chen2018fast,xu2019topology} to generate perturbations on graphs, which are then taken as input to the victim models.
However, directly applying these gray box attack methods to heterogeneous graphs does not work. 
Because: 1) these methods overlook heterogeneous information when designing the surrogate model, thus having difficulties in deriving gradient information within the message-passing paradigm; 
and 2) they treat all edges uniformly in the search for optimal perturbations, leading to the ignorance of the diverse semantic meanings associated with different relations, which will be elaborated as follows.

\textit{How to design an optimal surrogate model?} Due to the inherent complexity of heterogeneous graphs, it is suboptimal to directly apply homogeneous GNN to them by ignoring heterogeneous information~\cite{dong2017metapath2vec,wang2019heterogeneous}. Furthermore, while gradient-based methods have achieved promising results in gray box settings~\cite{chen2018fast,wu2019adversarial}, they need to acquire the gradient information of the adjacency matrix. However, the majority of HGNN models are formulated within a message-passing paradigm~\cite{wang2019heterogeneous,lv2021we}, which restricts gradient access to existing edges only, leaving gradients for non-existent edges unattainable. Compounding this issue is the substantial size of real-world heterogeneous graphs; directly computing the gradient of the entire adjacency matrix in such cases can lead to significant memory consumption. 

\textit{How to search for optimal perturbations among various relations within a constrained budget?} In HGs, a single target node might be connected to various nodes through multiple relations, each representing distinct semantic meanings and contributing unequally to the prediction results of the target node.
Adversarial attack methods for homogeneous graphs, such as FGA~\cite{chen2018fast}, typically select the edge with the highest absolute gradient value from all available edges. However, 
this approach is less effective for HGs as it 
treat all edges equally, disregarding the diverse semantic meanings associated with different relations.

In this paper, we focus on topology attacks and propose \textbf{HGAttack}, the first adversarial attack method
for heterogeneous graphs within the targeted gray box evasion attack scenario. Inspired by the prevalent gray box attack methods for homogeneous graphs, we design a novel surrogate model to closely resemble the behaviors of HGNNs and apply gradient-based methods for perturbation generation. Given that most HGNNs utilize aggregated meta-path information~\cite{wang2019heterogeneous,fu2020magnn,yang2023simple}, our surrogate model first extracts meta-path induced homogeneous graphs from the input heterogeneous graph. By aligning the operation of the surrogate model with the conventional approach in HGNNs, the transferability of our HGAttack on the target model is significantly enhanced. To leverage the advantage of gradient-based attack methods, our surrogate model then applies GNN models as the underlying learning framework to obtain node embeddings with distinct semantic information from each meta-path induced graph. This approach also aids in managing memory costs, by decomposing the global graph into multiple subgraphs based on meta-paths, when calculating the gradient information based on the downstream task loss. To identify optimal perturbations across different relations, HGAttack incorporates a semantics-aware mechanism that generates perturbations according to relations' contribution to the prediction. 

Extensive experiments on widely used datasets demonstrate the efficacy of HGAttack, as it diminishes the performance of target HGNN models notably, surpassing the degradation achieved by baselines.
We also empirically analyze the behavior of HGAttack, which shows that: 1) HGAttack tends to corrupt the original label distribution of the target nodes based on the different meta-path induced subgraphs; and
2) contrary to existing research~\cite{zhang2022robust} which posits large degree (hub) nodes will enlarge the perturbation effect, our results do not observe a pronounced correlation between these two. These insights contribute valuable perspectives to the ongoing development of robust HGNNs.

In summary, our contributions are: 
\begin{itemize}
    \item We propose HGAttack, the first gray box evasion attack method for Heterogeneous GNNs. 
    \item We develop a novel surrogate model which closely resembles the behavior of HGNNs, facilitating the generation of effective gray box attacks, while reducing the cost of calculating gradients. 
    \item We develop a semantics-aware mechanism to auto-generate gray box attacks by selecting the most optimal perturbations while adhering to a constrained budget.
    \item We conduct both quantitative experiments and qualitative analysis to show the effectiveness of HGAttack and the insights gained on developing robust HGNNs. 
\end{itemize}

\section{Related Work}

\subsection{Heterogeneous Graph Neural Networks}

Heterogeneous Graph Neural Networks (HGNNs) are proposed to extract rich heterogeneous information from graph data. Existing HGNNs can be classified into two categories. The first category includes approaches that perform information aggregation at various levels, guided by manually crafted meta-paths~\cite{dong2017metapath2vec,wang2019heterogeneous,yang2023simple}. For instance, MAGNN~\cite{fu2020magnn} first generates meta-path instances based on given meta-paths, and then performs intra-meta-path aggregation and inter-meta-path aggregation. 
CKD~\cite{wang2022collaborative} performs knowledge distillation among different meta-path induced graphs. 
The second category extends homogeneous GNN by incorporating type-specific aggregation over one-hop neighbors~\cite{hu2020heterogeneous,schlichtkrull2018modeling}. For instance, SimpleHGN~\cite{lv2021we} extends GAT~\cite{velivckovic2017graph} by introducing edge-type embeddings when calculating attention weights. 
\begin{figure*}[h]
    \centering
    \includegraphics[width=1.0\linewidth]{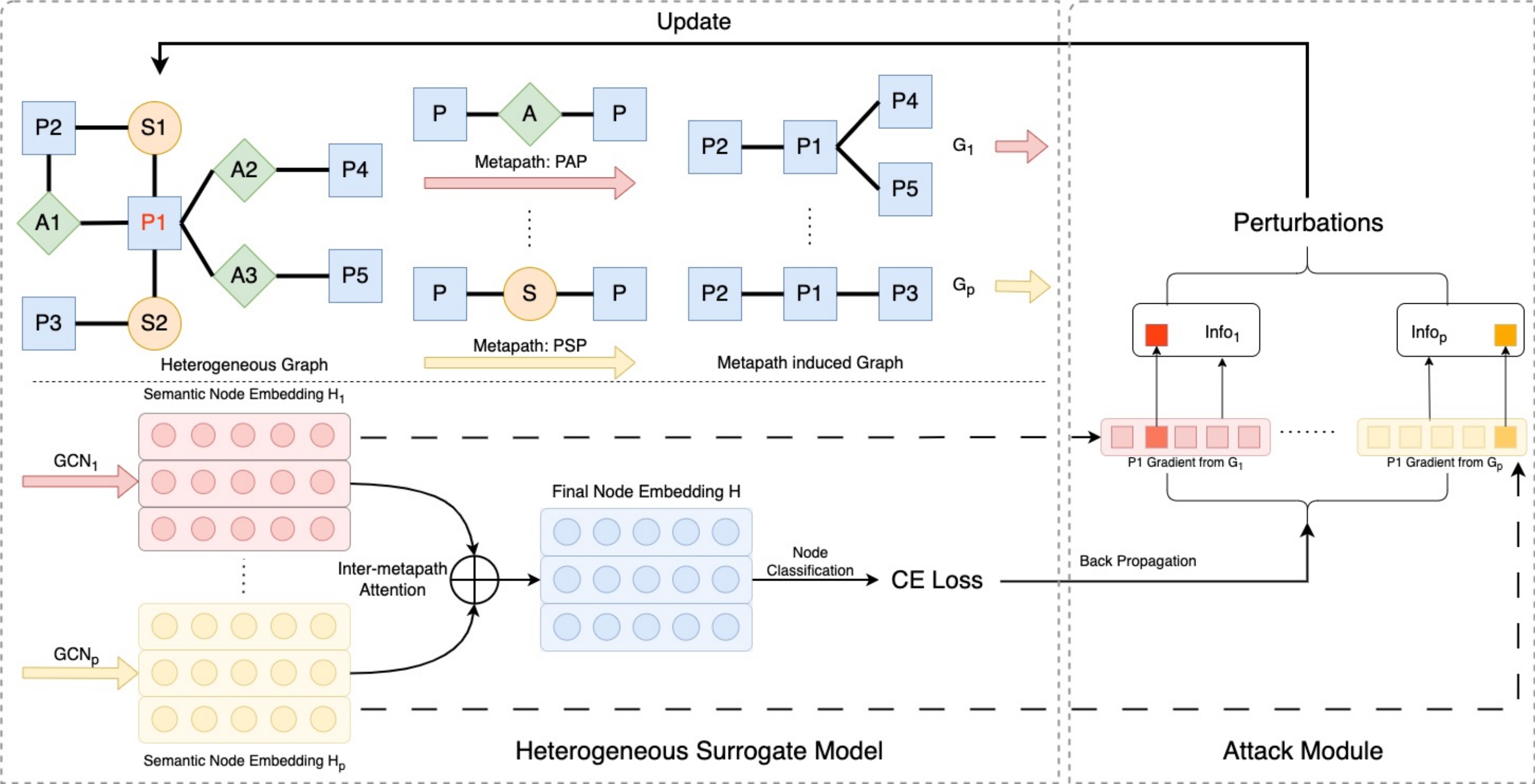}
    \caption{Overview of our HGAttack method. 
    Our heterogeneous surrogate model first decomposes the input graph into meta-path induced graphs. Different GCNs are applied to learn different semantic embeddings and fused into the final embeddings through inter-meta path attention. The attack module searches for perturbations via gradient information with our semantics-aware mechanism. The perturbations are used to update the input graph and then iteratively generate perturbations until it reaches the budget size.}
    \label{fig:model}
\end{figure*}
\subsection{Adversarial Attacks on Graph Data}

Recently, a number of studies have emerged to investigate adversarial attacks on graph data~\cite{zugner2018adversarial,dai2018adversarial,wu2019adversarial,li2022revisiting,sharma2023task}. Based on the attacking stage, attacks can be categorized into poisoning attacks~\cite{wang2019attacking} (happen in the training stage) and evasion attacks (happen in the testing stage)~\cite{dai2018adversarial}. Additionally, attacks are classified according to the attackers' knowledge into white box, gray box, and black box attacks~\cite{wei2020adversarial}. In the context of gray box evasion attacks, the attackers possess access solely to the training labels and adjacency matrices, lacking any knowledge about the targeted model. Typically, in such scenarios, attackers initially utilize a surrogate model such as GCN~\cite{kipf2016semi} to generate perturbations using white box attack methods, e.g.FGA~\cite{chen2018fast}, PGDattack~\cite{xu2019topology}. Despite the effectiveness of these methods in homogeneous graph contexts, as discussed in the introduction, their direct application to heterogeneous graphs is impeded by their limited ability to accommodate the structural and semantic complexity of HGNNs.

Recent research by~\cite{zhang2022robust} has pioneered the study of the robustness of HGNNs. They identified two key reasons for the vulnerabilities of HGNNs, the perturbation enlargement effect and the soft attention mechanisms, and methods to tackle them. However, their focus was primarily on defense mechanisms, with less attention on the attack method design. Their approach of generating perturbations, which employs GCN on one single selected meta-path-induced homogeneous graph, could not comprehensively capture the global heterogeneous information. Moreover, the need to pre-select a targeted relation restricts its capacity to identify the most optimal perturbation among the various available relations. We argue that perturbations generated through this approach may lack comprehensive potency, potentially leading to an incomplete evaluation of HGNNs’ robustness against adversarial attacks.

\section{Methodology}

\subsection{Preliminaries}
In this section, we give formal definitions of concepts and essential notations related to this work.

\textbf{Definition 1. Heterogeneous Graphs}. A heterogeneous graph is defined as $\mathcal{G}=\{\mathcal{V}, \mathcal{E}, X,\phi,\psi\}$, where $\mathcal{V}$ is a node set, $\mathcal{E}$ denotes an edge set and $X \in \mathbb{R} ^{\left|\mathcal{V}\right|\times f}$ represents a node feature matrix where $f$ denote the dimension of a node feature. $\phi$ and $\psi$ denote a node type mapping function $\phi: \mathcal{V}\rightarrow \mathcal{A}$ and an edge type mapping function $\psi: \mathcal{E} \rightarrow \mathcal{R}$, where $\mathcal{A}$ and $\mathcal{R}$ are predefined node type set and edge type set, respectively. A heterogeneous graph satisfies a restriction that $\left|\mathcal{A}\right|+\left|\mathcal{R}\right|>2$. Each relation $r \in \mathcal{R}$ is associated with a binary adjacency matrix $A_r$. 

\textbf{Definition 2. Meta-paths}. A meta-path is manually defined in the form of $r_1 \circ r_2 \circ \cdots \circ r_l$, where $\circ$ denotes a composite operation and $r_i \in \mathcal{R}$. 
Given a meta-path $P$, a meta-path induced homogeneous graph $G_P$ indicates that, for each edge in $G_P$, there exists at least one path based on $P$ between the head node and tail node. 

\textbf{Definition 3. Targeted Topology Evasion Attack.}
In the test phase, given a target node $v_i \in \mathcal{V}$ and a well-trained HGNN $\mathcal{F}_{HGNN}$, the goal of targeted topology evasion attack is to find the optimal edge perturbations constrained by a budget $\Delta$ to force $\mathcal{F}_{HGNN}$ make a wrong prediction on $v_i$. Formally, the perturbated adjacency matrix $\Tilde{A}_r$ is defined as follows:
\begin{equation}
        \Tilde{A}_r = A_r + C \circ S; C = \Bar{A}_r-A_r; ||S|| \leq \Delta,
\label{eq:adj}
\end{equation}
where $S_{ij}\in \{0,1\}$ is a Boolean perturbation indicating matrix, representing whether the corresponding edge is perturbated or not.  $C$ is the edge matrix that can be perturbated, the negative values represent the edges that can be deleted and the positive ones represent those that can be added. $\Bar{A}_r = \textbf{1}-A_r$ is the complementary matrix of $A_r$. 

\subsection{HGAttack: An Overview}
In this paper, we propose the first adversarial attack method named \textbf{HGAttack} for heterogeneous graphs within the targeted gray box evasion topology attack scenario. {To effectively extract comprehensive global heterogeneous information while optimizing memory consumption, we develop a novel surrogate model that can approximate the behavior of HGNNs and also facilitate the application of gradient-based attacking methods. To enable flexible selection of vulnerable relations, we also introduce a semantics-aware mechanism to generate perturbations. The overall working flow of our HGAttack is shown in Figure ~\ref{fig:model}.

\subsection{Heterogeneous Surrogate Model}
Homogeneous GNN fails to directly serve as the surrogate model for HGNNs because they cannot handle heterogeneity appropriately. To design an effective heterogeneous surrogate model, we need to address the following challenges: 1) \textbf{Transferability}. We anticipate that the perturbations generated by the surrogate model can perturb and adversely affect the predictions of the targeted model; 2) \textbf{Capacity}. HGs encompass a multitude of relations, each bearing distinct semantic significance. We anticipate that our attack methods can adeptly leverage the heterogeneity within HGs, allowing for the flexible selection of relations to attack, rather than necessitating the pre-definition of specific relations. 3) \textbf{Efficiency}. HGs typically encompass large-scale networks, making it computationally time-consuming to directly calculate gradients for the entire adjacency matrix. 

Recently, most HGNNs have adopted meta-paths to model the heterogeneity and perform information aggregation at different levels~\cite{dong2019learning,fu2020magnn,yang2023simple}. To enhance the transferability of the generated perturbations, it is essential for the surrogate model to closely resemble the targeted models in terms of their operational mechanisms~\cite{demontis2019adversarial}. To this end, we propose to utilize meta-path information to imitate the behavior of HGNNs. Given the meta-path set $\Phi$, we first extract multiple meta-path induced homogeneous graphs $A_{P}$, where $P\in \Phi$. Specifically, $A_{P}$ is defined as follows, 
\begin{align}
\begin{split}
    A_{P} &= A_{r_1}A_{r_2}\cdots A_{r_l}, \\
    P &= r_1 \circ r_2 \circ \cdots \circ r_l, P \in \Phi,
\end{split}
\end{align}
where $A_{r_1}$ is the binary adjacency matrix for relation $r$. It is important to note that in order to enable our surrogate model to comprehensively explore the heterogeneous information, we ensure that each selected meta-path corresponds to a unique relation, and the set of meta-paths collectively covers all available relations in an equivalent manner. Then we input these graphs into different GCNs~\cite{kipf2016semi} to obtain different semantic node embeddings $H_{p}$. For each GCN,  each layer is represented as follows, 
\begin{align}
    H_{p}^{(l)} &= \sigma(\hat{{A}}_PH_{p}^{(l-1)}W^{(l)}), \\
    \hat{A}_P &= {D_p}^{-\frac{1}{2}}(A_P+I)D_p^{-\frac{1}{2}},
\end{align}
where $D$ is the degree matrix, $W^{(l)}$ is the weight of the $l$th layer. In this way, we can employ gradient-based methods to calculate the gradients of $A_r$. Furthermore,  we decompose the overall heterogeneous graph into multiple subgraphs which also reduce the memory cost. To imitate the behavior of HGNNs, we then apply an inter-meta-path attention mechanism to fuse the different semantic embeddings. 
\begin{align}
    \begin{split}
        \alpha_{i} &= \frac{\mathrm{exp}(\mathrm{tanh}(a^T[\Tilde{W}H_i]))}
    {\sum_{j\in \Phi}\mathrm{exp}(\mathrm{tanh}(a^T[\Tilde{W}H_j]))}, \\
        H &= \sum_{i\in \Phi}\alpha_iH_i,
    \end{split}
\end{align}
where $\Tilde{W}$ is the weight matrix, $\alpha_i$ is the attention weight vector, $H$ is the final node embeddings. We apply the Cross-Entropy(CE) loss as our training objective for the surrogate model. 
\subsection{Generate Perturbations}
Given the trained surrogate model $\mathcal{F}_{HGNN}$ and the target node set $V_t$, we first obtain the pseudo labels $\hat{y}_i$ for each node $v_i \in V_t$. The pseudo labels are then used for calculating the loss. The attacking objective can be formulated as follows: 
\begin{align}
    \max_{||S||\leq \Delta}\mathcal{L}_{CE}(\mathcal{F}_{HGNN}(\Tilde{A}_1, \Tilde{A}_2,\cdots, \Tilde{A}_r, X)_i, \hat{y}_i),
\end{align}
where $S$ is the Boolean perturbation indicating matrix, $X$ is the node feature matrix, $\mathcal{F}_{HGNN}(\cdot)_i$ is the prediction of node $v_i$. The main idea is to maximize the loss to make the model make wrong predictions constrained by the perturbation budget. 
Note that we notice that in practice, the surrogate model may produce overconfidence predictions. In this case, the loss is 0 will lead to the gradients being 0.  Thus, we switch to the pseudo label to the class with the second large logit during prediction.
Then we propose a gradient-based method based on the FGA~\cite{chen2018fast} together with a newly developed semantics-aware mechanism to generate perturbations iteratively. In each interaction, given the loss of node $v_i$, the gradient of each binary adjacency is calculated via,
\begin{align}
    Grad_{rij} = (\frac{\partial L_{CE_{i}}}{\partial \hat{A}_{r}})_{ij},
\end{align}
Drawing inspiration from findings in homogeneous graph adversarial attacks, which highlight the potency of direct attacks (where one node on the perturbation edge is the target node) compared to influencer attacks (where the target node is not on the perturbation edge)~\cite{zugner2020adversarial}, we advocate for selecting perturbations from direct relations which are defined similarly. Since different relations correspond to different semantic meanings with different significance to the prediction results, we propose a semantics-aware mechanism to adjust the weight of the gradients. The value of weighted gradients is defined as follows,
\begin{align}
    \Hat{Grad}_{rij} = ||Grad_r)|| * Grad_{rij},
\end{align}
where $Grad_r$ is the gradient matrix of $\hat{A}_r$, we use the norm $||Grad_r||$ to represent the semantic significance of the relation $r$. Then we select the perturbations based on the absolute value of the weighted gradients. Finally, we update $\hat{A}_r$ and its corresponding reverse relation adjacency matrix since the HGs are undirected. The overall algorithm is defined as follows.
\begin{algorithm}
    \renewcommand{\algorithmicrequire}{\textbf{Input:}}
    \renewcommand{\algorithmicensure}{\textbf{Output:}}
    \caption{Generate Perturbations} 
	\label{alg} 
	\begin{algorithmic}[1]
            \REQUIRE Trained surrogate model $\mathcal{F}_{HGNN}$, target node $v_t$, pseudo labels $\hat{y}$, adjacency matrix $A_r$, perturbation budget $\Delta$
            \ENSURE $\hat{A}_r$

            \STATE Initialize: $S_{ij} \gets 0, \hat{A}_r \gets A_r$
            \FOR{$iter \in range(\Delta)$}
            \STATE $\mathcal{L}_{CE_{t}} = \mathcal{L}_{CE}(\mathcal{F}_{HGNN}(\Tilde{A}_1, \Tilde{A}_2,\cdots, \Tilde{A}_r, X)_t, \hat{y}_t)$
            \STATE $Grad_{r} = (\frac{\partial L_{CE_{t}}}{\partial \hat{A}_{r}}), r\in \mathcal{R}$
            \STATE $\Hat{Grad}_{rtj}\! =\! ||Grad_r||*Grad_{rtj}, j\in range(len(A_{r_t}))$
            \vspace{-0.1in}
            \STATE $i = \argmax_{j}\{|\Hat{Grad}_{rtj}|\}$
            \STATE $S_{ti} = 1$
            \STATE Update $\hat{A}_r$ according to eq \ref{eq:adj} using $S$
            \STATE Update reverse relation $\hat{A}_r^T$ according to eq \ref{eq:adj} using $S^T$
            \ENDFOR
            
        \end{algorithmic}
\label{alg:hgattack}
\end{algorithm}

\section{Experiments}
\subsection{Experimental Settings}
We conduct experiments on three heterogeneous graph benchmark datasets,which are ACM, DBLP and IMDB. Details are summarized in the Table ~\ref{tab:dataset}.
We randomly select 500 nodes as the targeted nodes from the original test set used for the training of the surrogate model of each dataset. We set the GCN used in the surrogate model to two layers. We select 3 classic HGNN models, i.e. HAN~\cite{wang2019heterogeneous}, HGT~\cite{hu2020heterogeneous},  and 1 robust HGNN model RoHe~\cite{zhang2022robust}. For each target node, we constrain the perturbation budget $\Delta\in\{1,3,5\}$. Each victim model is well-trained first. We reload the original graph and update it to the perturbated graph based on the attacking results corresponding to the target node at each test time. For each target model, we compare our HGAttack with the homogeneous attack method FGA~\cite{chen2018fast} and the heterogeneous attack method proposed in ~\cite{zhang2022robust}, henceforth referred to as the HG Baseline. We choose GCN as the surrogate model of the FGA method by ignoring the heterogeneous information. As the generation of perturbations necessitates gradient information from the adjacency matrix, our implementation of the GCN is based on the spectral GNN framework~\cite{wang2022powerful}, diverging from the message-passing paradigm~\cite{velivckovic2017graph}.


\begin{table}[]
\centering
\small
\renewcommand{\arraystretch}{1.0}
\setlength{\tabcolsep}{1.4mm}{
\begin{tabular}{|l|l|rrr|r|}
\hline
Dataset               & Relations(A-B) & \# of A & \# of B & \# of A-B & MPs   \\ \hline
\multirow{2}{*}{ACM}  & Paper-Author   & 3025    & 5912    & 9936      & PAP   \\
                      & Paper-Subject  & 3025    & 57      & 3025      & PSP   \\ \hline
\multirow{2}{*}{IMDB} & Movie-Director & 4661    & 2270    & 4661      & MDM   \\
                      & Movie-Actor    & 4661    & 5841    & 13983     & MAM   \\ \hline
\multirow{3}{*}{DBLP} & Author-Paper   & 4057    & 14328   & 19645     & APA   \\
                      & Paper-Venue    & 14328   & 20      & 14328     & APVPA \\
                      & Paper-Term     & 14328   & 7723    & 85810     & APTPA \\ \hline
\end{tabular}%
}
\caption{Statistics of the datasets and the meta-paths used in our experiments. Each meta-path has a corresponding relation, of which the corresponding binary adjacency matrix serves as the trainable matrix to obtain the gradient when generating perturbations.}
\label{tab:dataset}
\end{table}

\begin{table*}[t]
\centering
\renewcommand{\arraystretch}{1.0}
\resizebox{\textwidth}{!}{%
\begin{tabular}{|l|l|l|cc|cc|cc|cc|}
\hline
\multirow{2}{*}{Dataset} & \multirow{2}{*}{Target Model} & \multirow{2}{*}{Attack Method} & \multicolumn{2}{c|}{Clean}                   & \multicolumn{2}{c|}{$\Delta=1$} & \multicolumn{2}{c|}{$\Delta=3$} & \multicolumn{2}{c|}{$\Delta=5$} \\ \cline{4-11} 
                         &                               &                                & Macro F$_1$             & Micro F$_1$              & Macro F$_1$       & Micro F$_1$       & Macro F$_1$       & Micro F$_1$       & Macro F$_1$       & Micro F$_1$       \\ \hline
\multirow{12}{*}{ACM}    & \multirow{3}{*}{HAN}          & FGA                            & \multicolumn{1}{l}{} & \multicolumn{1}{l|}{} & 76.32          & 76.20          & 51.47          & 50.40          & 44.47          & 43.4           \\
                         &                               & HG Baseline                    & 91.37                & 91.20                 & 60.57          & 60.60          & 51.61          & 52.20          & 49.24          & 50.2           \\
                         &                               & HGAttack                       &                      &                       & \textbf{16.12} & \textbf{16.20} & \textbf{11.31} & \textbf{13.80} & \textbf{18.40} & \textbf{19.60} \\ \cline{2-11} 
                         & \multirow{3}{*}{HGT}          & FGA                            & \multicolumn{1}{l}{} & \multicolumn{1}{l|}{} & 89.78          & 89.60          & 79.90          & 79.60          & 76.41          & 76.20          \\
                         &                               & HG Baseline                    & 92.31                & 92.20                 & 84.20          & 84.20          & 79.19          & 79.20          & 74.96          & 75.00          \\
                         &                               & HGAttack                       &                      &                       & \textbf{79.16} & \textbf{79.40} & \textbf{66.69} & \textbf{66.80} & \textbf{58.18} & \textbf{57.80} \\ \cline{2-11} 
                         & \multirow{3}{*}{SimpleHGN}    & FGA                            & \multicolumn{1}{l}{} & \multicolumn{1}{l|}{} & 75.12          & 75.00          & 39.14          & 39.60          & 26.89          & 27.60          \\
                         &                               & HG Baseline                    & 90.37                & 90.20                 & 72.18          & 72.00          & 54.16          & 54.20          & 57.60          & 57.60          \\
                         &                               & HGAttack                       &                      &                       & \textbf{66.23} & \textbf{65.40} & \textbf{25.64} & \textbf{22.60} & \textbf{13.38} & \textbf{10.40} \\ \cline{2-11} 
                         & \multirow{3}{*}{RoHe}         & FGA                            & \multicolumn{1}{l}{} & \multicolumn{1}{l|}{} & \textbf{81.31} & \textbf{81.00} & \textbf{76.25} & \textbf{76.00} & \textbf{85.11} & \textbf{84.80} \\
                         &                               & HG Baseline                    & 91.19                & 91.00                 & 90.54          & 90.40          & 90.55          & 90.40          & 90.18          & 90.00          \\
                         &                               & HGAttack                       &                      &                       & 90.77          & 90.60          & 89.99          & 89.80          & 89.63          & 89.40          \\ \hline
\multirow{12}{*}{IMDB}   & \multirow{3}{*}{HAN}          & FGA                            & \multicolumn{1}{l}{} & \multicolumn{1}{l|}{} & 57.62          & 59.40          & 54.32          & 55.60          & 47.09          & 47.20          \\
                         &                               & HG Baseline                    & 57.28                & 59.00                 & 32.86          & 33.80          & 27.73          & 28.80          & 29.94          & 31.20          \\
                         &                               & HGAttack                       &                      &                       & \textbf{27.96} & \textbf{27.60} & \textbf{12.33} & \textbf{11.80} & \textbf{11.73} & \textbf{11.60} \\ \cline{2-11} 
                         & \multirow{3}{*}{HGT}          & FGA                            & \multicolumn{1}{l}{} & \multicolumn{1}{l|}{} & 58.30          & 61.80          & 55.85          & 58.00          & 57.57          & 61.20          \\
                         &                               & HG Baseline                    & 58.40                & 61.00                 & 27.08          & 27.40          & 21.00          & 22.40          & 24.18          & 24.00          \\
                         &                               & HGAttack                       &                      &                       & \textbf{25.90} & \textbf{25.20} & \textbf{15.16} & \textbf{14.20} & \textbf{14.74} & \textbf{14.20} \\ \cline{2-11} 
                         & \multirow{3}{*}{SimpleHGN}    & FGA                            &                      &                       & 52.57          & 55.80          & 45.61          & 47.80          & 43.90          & 44.20          \\
                         &                               & HG Baseline                    & 57.36                & 59.80                 & 29.67          & 29.40          & 26.81          & 26.80          & 35.40          & 36.60          \\
                         &                               & HGAttack                       &                      &                       & \textbf{28.81} & \textbf{28.00} & \textbf{15.21} & \textbf{14.40} & \textbf{13.00} & \textbf{12.20} \\ \cline{2-11} 
                         & \multirow{3}{*}{RoHe}         & FGA                            &                      &                       & 50.24          & 51.00          & 50.58          & 51.20          & 48.85          & 49.20          \\
                         &                               & HG Baseline                    & 52.64                & 49.20                 & 51.08          & 51.40          & 50.16          & 50.40          & 49.28          & 49.80          \\
                         &                               & HGAttack                       &                      &                       & \textbf{48.68} & \textbf{49.20} & \textbf{44.97} & \textbf{44.50} & \textbf{42.52} & \textbf{42.40} \\ \hline
\multirow{12}{*}{DBLP}   & \multirow{3}{*}{HAN}          & FGA                            &                      &                       & --             & --             & --             & --             & --             & --             \\
                         &                               & HG Baseline                    & 92.28                & 93.20                 & 52.89          & 54.80          & 23.74          & 27.80          & 18.11          & 18.20          \\
                         &                               & HGAttack                       &                      &                       & \textbf{36.61} & \textbf{38.20} & \textbf{10.58} & \textbf{10.80} & \textbf{10.05} & \textbf{10.40} \\ \cline{2-11} 
                         & \multirow{3}{*}{HGT}          & FGA                            &                      &                       & --             & --             & --             & --             & --             & --             \\
                         &                               & HG Baseline                    & 93.28                & 94.00                 & 69.53          & 70.00          & 44.23          & 45.40          & 40.48          & 41.00          \\
                         &                               & HGAttack                       &                      &                       & \textbf{67.20} & \textbf{69.40} & \textbf{34.85} & \textbf{36.00} & \textbf{25.18} & \textbf{25.40} \\ \cline{2-11} 
                         & \multirow{3}{*}{SimpleHGN}    & FGA                            &                      &                       & --             & --             & --             & --             & --             & --             \\
                         &                               & HG Baseline                    & 93.43                & 94.20                 & 81.20          & 82.40          & 44.22          & 45.22          & 39.24          & 39.60          \\
                         &                               & HGAttack                       &                      &                       & \textbf{68.98} & \textbf{71.00} & \textbf{25.59} & \textbf{25.60} & \textbf{17.78} & \textbf{17.80} \\ \cline{2-11} 
                         & \multirow{3}{*}{RoHe}         & FGA                            &                      &                       & --             & --             & --             & --             & --             & --             \\
                         &                               & HG Baseline                    & 91.78                & 92.60                 & 91.98          & 92.80          & 91.79          & 92.60          & 91.34          & 92.20          \\
                         &                               & HGAttack                       &                      &                       & \textbf{90.71} & \textbf{91.60} & \textbf{87.22} & \textbf{88.00} & \textbf{83.58} & \textbf{84.59} \\ \hline
\end{tabular}%
}

\caption{Overall attack results for the three datasets. Lower scores mean more powerful attacking ability. ``\textbf{Clean}'' means the original graphs without adversarial perturbations. ``--'' means the method fails to run on a 16 GB GPU due to the memory cost.}
\label{tab:results}
\end{table*}

\subsection{Attacking Results}
\label{sec:main_res}
We report the average scores over the selected 500 target nodes on each dataset. Results are summarized in the Tables ~\ref{tab:results}. 
Across all target models and datasets evaluated, our proposed HGAttack markedly degrades the performance of the target HGNN models, consistently surpassing the level of degradation achieved by baseline methods significantly. The only exception is observed with the RoHe model on the ACM dataset. In this scenario, the FGA method, which typically exhibits poor performance across other cases due to its disregard for heterogeneous information, unexpectedly shows a strong attacking ability.
This unusual case can be attributed to two factors. Firstly, FGA performs better in corrupting the original label distributions on the ACM dataset than on the IMDB dataset (see Section~\ref{sec:analysis} for detailed explanations), hence it degrades the performance of target models more effectively on the ACM dataset. Secondly, the defense mechanisms of RoHe render it particularly vulnerable to low-degree adversarial nodes. RoHe is designed to mitigate the `hub' effect of large degree nodes, assigning them low scores in contrast to high scores for low degree nodes. We observe that the FGA tends to generate more low degree adversarial nodes than the HG Baseline and our HGAttack. The convergence of these two factors results in RoHe's vulnerability to FGA in the context of the ACM dataset.

Besides, we observe that an increase in the perturbation budget size does not invariably enhance the attack's efficacy; for instance, the capability to compromise the HAN model diminishes when the perturbation budget is escalated from $\Delta=3$ to $\Delta=5$. This counterintuitive outcome may be ascribed to potential overfitting on the surrogate model. A lower perturbation budget at this inflection point reflects a less potent attacking ability, further corroborating the superiority of our method over the baseline approach.

\begin{figure*}[t]
\centering
\subfloat[ACM]{
\includegraphics[width=0.3\linewidth]{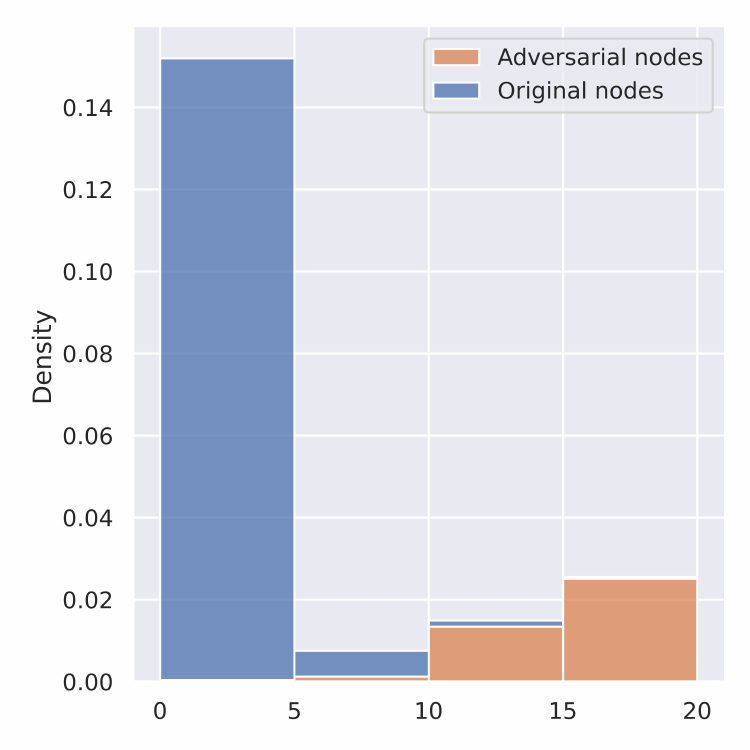}
}
\subfloat[IMDB]{
\includegraphics[width=0.3\linewidth]{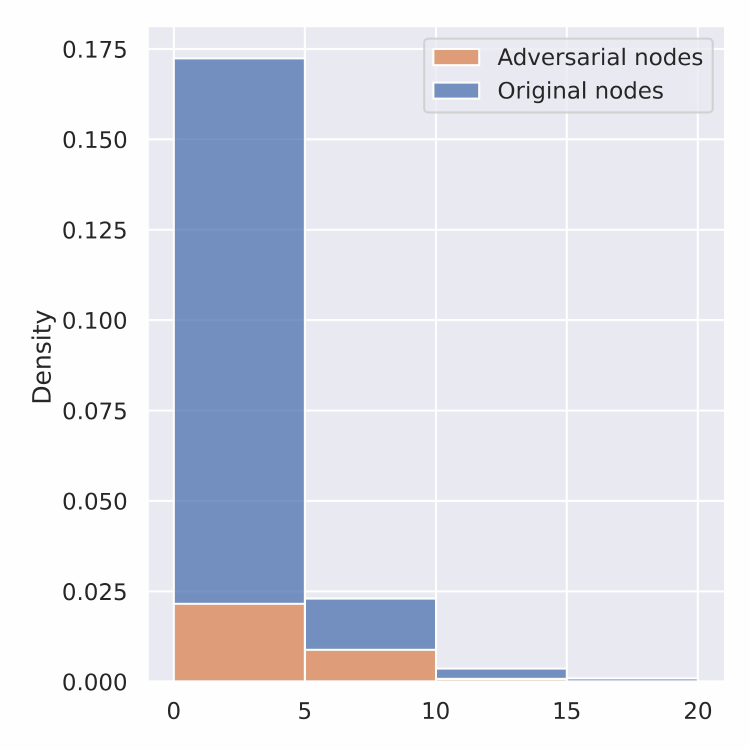}
}
\subfloat[DBLP]{
\includegraphics[width=0.3\linewidth]{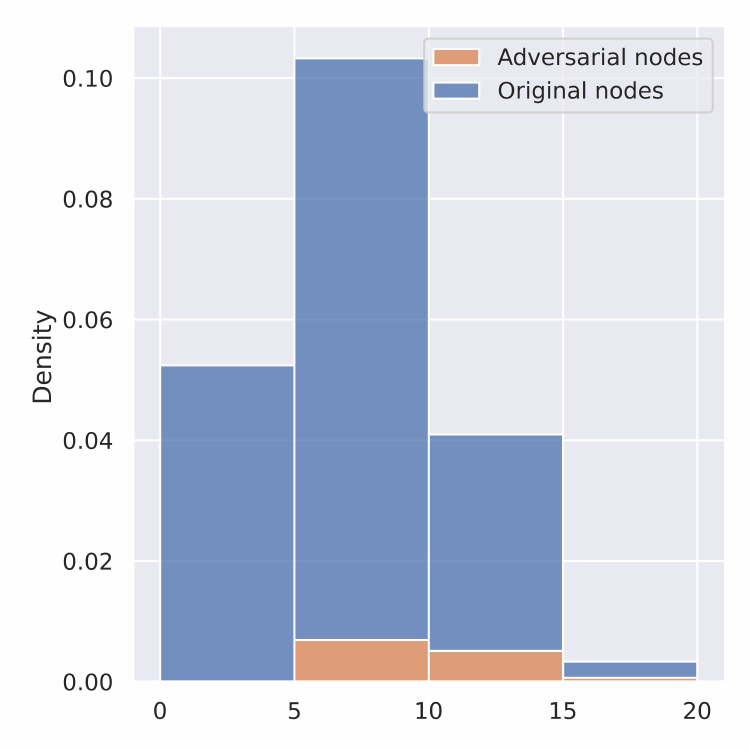}
}

\caption{Histograms illustrating degree distributions of neighbor-type nodes and adversarial nodes across three datasets, with node degree on the X-axis and density on the Y-axis.}
\label{fig:degree_dis}
\end{figure*}

Upon further examination, we note that among the three HGNN models evaluated, HAN exhibits the least robustness, evidenced by the most significant performance decline when subjected to identical perturbations. This vulnerability can be attributed to HAN's utilization of meta-path aggregation, as different from the one-hop neighbor aggregation employed by the other models. The method of aggregation in HAN potentially facilitates susceptibility to attacks, due to an amplification of perturbations, a phenomenon described by~\cite{zhang2022robust}. Conversely, this perturbation magnification effect appears to be mitigated in HGNN models that implement one-hop neighbor aggregation. RoHe \cite{zhang2022robust} shows its robustness by filtering out low confidence score neighbors based on the feature similarity and transition score.

\subsection{Emperical Analysis}\label{sec:analysis}

We further perform empirical analysis to study the behavior of our HGAttack. 
In the existing study~\cite{zhang2022robust}, the authors propose a perturbation enlargement effect that the large degree node will lead to an increasing number of adversarial meta-path instances. They also conclude through their experiments that attack methods tend to generate perturbations between large degree nodes and the target nodes. To verify this conclusion, we compare the neighbor-type node (nodes connected to the target-type nodes) degrees and the adversarial node degrees in the original graphs. The node distributions are shown in the Figure \ref{fig:degree_dis}. 

For the ACM dataset, we observe that the distribution of adversarial nodes, unlike that of neighbor-type nodes, predominantly centers around large-degree nodes. This trend aligns with the conclusions drawn in~\cite{zhang2022robust}. However, for IMDB and DBLP datasets, the distribution of the adversarial nodes is more similar to that of the neighbor-type nodes. There is no obvious tendency for HGAttack to link large degree nodes to the target nodes. To provide a more accurate evaluation, we calculated the number of adversarial nodes exceeding the 90th percentile of original node degrees in Table~\ref{tab:degree_dis}. The statistics reveal a significant disparity across the three datasets. Consequently, we posit that there is no clear inclination for heterogeneous graph attack methods to generate perturbations between large degree nodes and target nodes. Furthermore, the efficacy of such large degree node in corrupting HGNNs is also questionable.

From another perspective, we further investigate the relationship between the labels of the target nodes and their adversarial meta-path-based neighbors. To quantify the relationship, we define a \textbf{label inconsistency score} for each target-type node on meta-path $P$, which is defined as follows, 
\begin{align}
    Score_{v_t}^P &= \frac{1}{|\mathcal{N}^P(v_t)|}\sum_{v_i\in \mathcal{N}^P(v_t)}{\delta_i}, \\
    \delta_i &= \frac{1}{|\Tilde{\mathcal{N}}^P(v_i)|}\sum_{v_j \in \Tilde{N}^P(v_i)}{\mathbf{I}(y_t \neq y_j)},
\end{align}
where $\mathcal{N}(v_t)$ is the one-hop neighbors for the target node based on the meta-path $P$, $\Tilde{\mathcal{N}}(v_i)$ is the two-hop neighbors for the target node based on the meta-path $P$, $\mathbf{I}(y_t \neq y_j)$ equals to 1 if the two-hop neighbor node label is different from the target node label, otherwise it equals to 0. The overall score is the mean value of the $Score_{v_t}$ for all the target nodes. We compare the scores of the clean graph and the subgraphs generated by the perturbations with $\Delta=3$ using different attack methods in Tabel ~\ref{tab:label}.

\begin{table}[]
    \centering
    \renewcommand{\arraystretch}{1.0}
    \resizebox{\columnwidth}{!}{%
    \begin{tabular}{l|c|c|c}
\hline
                                 & ACM   & IMDB  & DBLP  \\ \hline
Large Degree Thresholds           & 3     & 5     & 11    \\
Percentages of LDA Nodes(\%) & 99.93 & 45.73 & 18.63 \\ \hline
\end{tabular}%
}
\caption{Statistical analysis of large degree adversarial nodes generated by HGAttack, with large degree defined as the 90th percentile of original node degrees. ``LDA Nodes'' refers to Large Degree Adversarial Nodes.}
    \label{tab:degree_dis}
\end{table}

\begin{table}[]
\centering
\small
\renewcommand{\arraystretch}{1.0}
\resizebox{\columnwidth}{!}{%
\begin{tabular}{|l|cc|cc|c|}
\hline
Methods     & \multicolumn{2}{c|}{ACM}                                 & \multicolumn{2}{c|}{IMDB}                                & DBLP                        \\ \hline
Meta-path   & \multicolumn{1}{c}{PAP}    & PSP                         & \multicolumn{1}{c}{MDM}    & MAM                         & \multicolumn{1}{c|}{APA}    \\ \hline
No Attack   & 0.0703                     & 0.3388                      & 0.1088                     & 0.3683                      & 0.0114                      \\ \hline
FGA         & 0.7026                     & \multicolumn{1}{c|}{0.0087} & 0.1193                     & \multicolumn{1}{c|}{0.2686} & --                          \\
HG Baseline & \multicolumn{1}{c}{0.7248} & *                           & \multicolumn{1}{c}{0.6192} & *                           & \multicolumn{1}{c|}{0.6120} \\
HGAttack    & 0.8969                     & 0.0228                      & 0.5873                     & 0.1856                      & 0.5940                      \\ \hline
\end{tabular}%
}
\caption{Label inconsistency score for the clean graphs and the adversarial perturbations induced subgraphs generated by different attack methods. For HG Baseline, we only report the results based on the selected meta-path used in the Table ~\ref{tab:results}.}
\label{tab:label}
\end{table}
We observe that the label distributions of the perturbated subgraphs generated by the two heterogeneous attack methods consistently deviate from those of the original graphs. Specifically, perturbated subgraphs for original graphs with high label inconsistency scores exhibit low inconsistency scores, and vice versa. Based on this observation, we posit that the effectiveness of these attack methods in degrading model performance stems from their ability to corrupt the original label distribution patterns. For the homogeneous attack method FGA, it effectively corrupted the label distribution in the ACM dataset, but did not manage to significantly alter the label distribution in the IMDB dataset. This is consistent with the results in Table~\ref{tab:results}, in which FGA shows relatively good attacking ability on the ACM dataset, but poor attacking performance on the IMDB dataset.
Such insights could be instrumental in formulating more effective strategies for both attacking and defending HGNN models.

\subsection{Ablation Studies}
In this part, we validate our semantics-aware mechanism. We propose a variant of our HGAttack named $\mathrm{HGAttack_{info}}$ which directly selects perturbations based on the gradient values, i.e. $\Hat{Grad}_{rtj} = Grad_{rtj}$ in line 5 in Algorithm~\ref{alg:hgattack}. Results are summarized in the Tabel ~\ref{tab:ablation}.
\begin{table}[]
\centering
\small
\renewcommand{\arraystretch}{1.0}
\resizebox{\columnwidth}{!}{
\begin{tabular}{|l|l|rr|rr|rr|}
\hline
\multirow{2}{*}{Dataset} & \multicolumn{1}{c|}{\multirow{2}{*}{Method}} & \multicolumn{2}{r|}{$\Delta=1$} & \multicolumn{2}{r|}{$\Delta=3$} & \multicolumn{2}{r|}{$\Delta=5$} \\ \cline{3-8} 
                         & \multicolumn{1}{c|}{}                        & Ma         & Mi                 & Ma         & Mi                 & Ma         & Mi                 \\ \hline
\multirow{3}{*}{IMDB}    & Baseline                                     & 27.08      & 27.40              & 21.00        & 22.40               & 24.18      & 24.40               \\
                         & $\mathrm{HGAttack_{info}}$                   & \textbf{19.22}      & \textbf{18.40}     & 19.73      & 18.40              & 25.17      & 24.00              \\
                         & HGAttack                                     & 25.90      & 25.20               & \textbf{15.16}      & \textbf{14.20}     & \textbf{14.74}      & \textbf{14.20}     \\ \hline
\multirow{3}{*}{DBLP}    & Baseline                                     & 69.53      & 70.00              & 44.23      & 45.40              & 40.48      & 41.00              \\
                         & $\mathrm{HGAttack_{info}}$                   & 69.58      & 71.40              & 35.09      & 36.40               & \textbf{25.15}      & \textbf{24.40}     \\
                         & HGAttack                                     & \textbf{67.20}      & \textbf{69.40}     & \textbf{34.85}      & \textbf{36.00}     & 25.18      & 25.40               \\ \hline
\end{tabular}%
}
\caption{Ablation study on the semantics-aware mechanism on the IMDB and DBLP datasets. Ma and Mi refer to the Macro F$_1$ score and the Micro F$_1$ score.}
\label{tab:ablation}
\end{table}

In general, HGAttack still achieves the best score compared with the baseline and $\mathrm{HGAttack_{info}}$. This observation suggests that while direct gradient-based selection is effective, it does not fully capitalize on the intricate relational dynamics present in heterogeneous graphs and further verify the effectiveness of our semantics-aware mechanism.

\subsection{Efficiency}
\begin{table}[]
\centering
\renewcommand{\arraystretch}{1.0}
\begin{tabular}{l|ccc}
\hline
            & ACM  & IMDB & DBLP \\ \hline
FGA & 4435 & 7573 & --   \\
HGAttack    & 2737 & 4247 & 7773 \\ \hline
\end{tabular}%
\caption{Memory cost (MB) comparison between our HGAttack and FGA on the three datasets.}
\label{tab:memory}
\end{table}

In this section, we verify the efficiency of our HGAttack. We conduct a comparative analysis between the homogeneous method FGA and our HGAttack. Results shown in the Table \ref{tab:memory} demonstrate that by decomposing the input graph into multiple meta-path induced homogeneous graphs, our HGAttack significantly reduces memory costs compared to traditional homogeneous graph methods. This finding further substantiates the efficiency of our proposed design.

\section{Conclusion}
In this paper, we introduce a method for heterogeneous graphs in the context of targeted gray box evasion attacks, called HGAttack. 
We develop a novel and effective surrogate model that harnesses the global information inherent in heterogeneous structures. 
We also develop a semantics-aware mechanism for identifying vulnerable edges across a range of relations within the constrained perturbation budget. 
Extensive experiments on benchmark datasets demonstrate the efficacy of HGAttack in executing targeted adversarial attacks. 
Empirical analysis on the attack tendency of HGAttack has also been conducted to help facilitate the design of robust HGNNs.
\bibliographystyle{named}
\bibliography{ijcai23}

\end{document}